\documentclass[letterpaper, 10 pt, conference]{ieeeconf}  

\usepackage{amsmath}
\IEEEoverridecommandlockouts                              

\usepackage{graphics} 
\usepackage{epsfig} 
\usepackage{times} 
\usepackage{amssymb}  
\usepackage{booktabs}
\usepackage{float}
\usepackage{booktabs}
\usepackage{booktabs}
\usepackage{times}
\usepackage{multirow} 
\usepackage{wrapfig}
\usepackage{cuted}
\usepackage{capt-of} 
\usepackage[caption=false,font=footnotesize]{subfig}
\usepackage{booktabs, tabularx, makecell}

\makeatletter
\let\NAT@parse\undefined
\makeatother
\usepackage[numbers,sort&compress]{natbib}

\usepackage{amsmath}

\title{\LARGE \bf
GALA: Geometry-Aware Latent Action Modeling for Vision-Language-Action Model Pretraining across Embodiments
}

\author{
Yichen Liu$^{1*}$ \quad
Puzhen Yuan$^{1*}$ \quad
Xiang Zhu$^{1,2*}$ \quad
Yanjiang Guo$^{1,2}$ \quad
Jianyu Chen$^{1,2\dagger}$ \\
$^{1}$ Institute for Interdisciplinary Information Sciences, Tsinghua University, China\\
$^{2}$ Shanghai Qi Zhi Institute, China\\
\texttt{\{liu-yc22, ypz21, zhuxiang24, guoyj22\}@mails.tsinghua.edu.cn} \\
\texttt{jianyuchen@mail.tsinghua.edu.cn} \\
$^{*}$Equal contribution, $^{\dagger}$Corresponding author.
}

\begin{document}

\maketitle
\thispagestyle{empty}
\pagestyle{empty}

\begin{figure*}[h]
    \centering
    \includegraphics[width=0.9\linewidth]{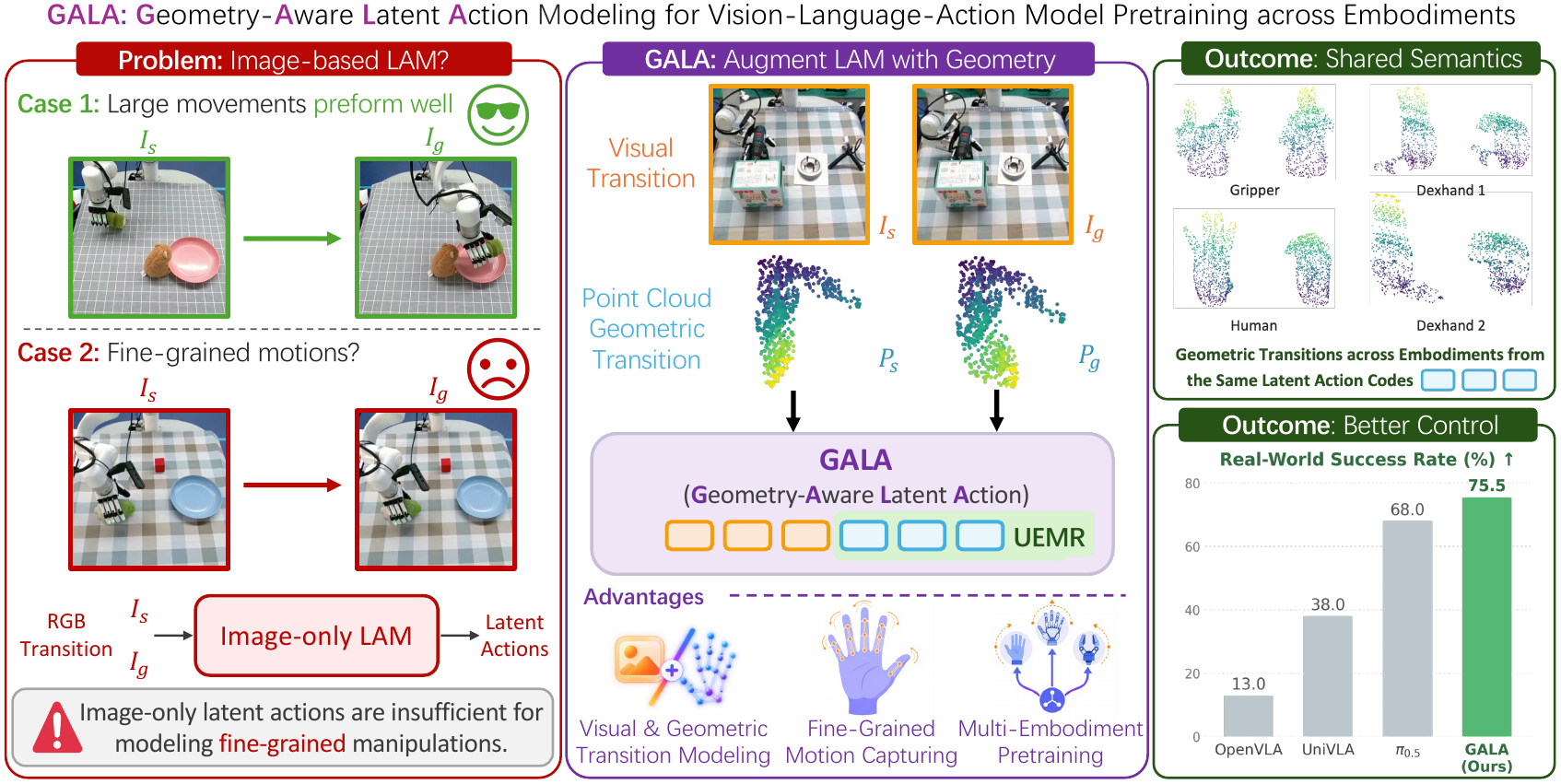}
    \vspace*{-3mm}
    \captionof{figure}{\textbf{Overview of GALA}. Geometry complements image-based latent actions with generalizable fine-grained end-effector motion, enabling more shared manipulation semantics across heterogeneous end-effectors, therefore enhancing VLA pretraining performance across embodiments.}
    \label{fig:teaser}
    \vspace*{-6mm}
\end{figure*}

\begin{abstract}

Learning large-scale vision-language-action (VLA) models from multi-embodiment datasets remains challenging due to heterogeneous action spaces across end effectors.
Although latent action models (LAMs) can learn embodiment-agnostic action representations from diverse video data, existing image-based LAMs often fail to capture fine-grained end-effector articulation, particularly finger-level geometric changes in human and dexterous robot hands.
To address this limitation, we propose GALA, a Geometry-Aware Latent-Action modeling framework that augments image-based latent actions with 3D end-effector geometric motion.
However, naively incorporating point clouds yields fine-grained action representations with limited shared semantics, hindering cross-embodiment pretraining. To address this issue, we introduce the Unified End-effector Motion Representation (UEMR), which preserves fine-grained motion information while improving the cross-embodiment generalizability of latent actions.
Building upon UEMR, GALA combines visual latent actions that capture scene-level dynamics with geometric latent actions that capture shared fine-grained end-effector articulation, providing effective supervision for VLA pretraining from multi-embodiment data, including action-free ego-centric human videos.
Experiments on fine-grained motion probing, cross-embodiment retrieval, and downstream VLA evaluation demonstrate GALA’s effectiveness in modeling generalizable fine-grained motions across embodiments, achieving 68.3\% RoboCasa-GR1 success rate and 75.5\% real-world success rate. 
Code, appendix, and demos are available at https://puzhenyuan.github.io/GALA-website/.

\end{abstract}


\section{Introduction}



Vision-language-action (VLA) models benefit from scaling manipulation data across tasks and embodiments ~\cite{Kim2024OpenVLAAO, kim2025fine, bjorck2025gr00t, Black20240AV, intelligence2025pi05, zhang2025up}, yet collecting large-scale robot demonstrations remains expensive.
Human videos provide a complementary source of diverse and dexterous interactions \cite{hu2024video, Fu2025METISME, Bauer2025LatentAD}, motivating joint learning from human and heterogeneous robot data. 
A key obstacle, however, is the mismatch between their action spaces: embodiments differ in kinematics, degrees of freedom, and control parameterizations, making native actions difficult to share without embodiment-specific alignment or retargeting~\cite{Fu2025METISME, Bauer2025LatentAD, Mu2026OnePolicyFitsAllGA, Jiang2026CrossHandLR}.


Latent action models (LAMs) offer a promising alternative by inferring action representations directly from visual transitions, enabling shared supervision from heterogeneous robot and action-free human videos~\cite{Ye2024LatentAP, Bu2025UniVLALT, zhu2026harp, Dai2026ConLACL}. 
However, existing image-based LAMs primarily capture scene-level end-effector displacement and may overlook fine-grained finger movements, particularly when motions are small or occluded, or viewpoints vary.
Consequently, they often capture \textbf{where} the end-effector moves but insufficiently characterize \textbf{how} it manipulates an object.


We propose \textbf{GALA}, a \textbf{G}eometry-\textbf{A}ware \textbf{L}atent \textbf{A}ction modeling framework that augments image-based latent actions with 3D end-effector geometric motion. 
By representing end-effector motions as point-cloud transitions, GALA explicitly captures fine-grained articulation and configuration changes without relying on native control representations. 
However, directly sharing geometry across embodiments introduces another challenge: geometric representations may encode morphology- and coordinate-specific cues rather than the underlying motion, resulting in latents that are not shared across heterogeneous hands and grippers.


To address this, we introduce a \textbf{U}nified \textbf{E}nd-effector \textbf{M}otion \textbf{R}epresentation (UEMR), which includes three core designs: unified bimanual motion latent, pair-consistent geometric augmentation, and bidirectional transition learning.  
Building upon the designs of UEMR, GALA learns shared geometric latents across human hands, dexterous robot hands, and parallel-jaw grippers without requiring point, joint, or kinematic correspondence. 
GALA combines these geometric latent actions with visual latent actions as complementary supervision for VLA pretraining, capturing both scene-level displacement and fine-grained end-effector articulation while retaining each embodiment's native executable action space.



We evaluate GALA through fine-grained motion probing, cross-embodiment retrieval, and downstream VLA learning in Robocasa~\cite{Nasiriany2024RoboCasaLS} and real-world settings. 
GALA preserves more fine-grained motion information, learns more transferable cross-embodiment action representations, and improves downstream VLA performance under heterogeneous embodiment co-training.

Our contributions are summarized as follows:

\begin{itemize}
\item We propose GALA, a geometry-aware latent-action modeling framework that augments image-based latent actions with 3D end-effector geometric motion, providing complementary scene-level and articulation-level supervision for VLA pretraining across embodiments.
\item We introduce UEMR, a unified end-effector motion representation tailored to GALA, which preserves fine-grained motion information while improving the cross-embodiment generalizability of latent actions.
\item Experiments on latent-action representation and downstream policy learning demonstrate the effectiveness of GALA in RoboCasa and real-world environments.
\end{itemize}

\section{Related Work}

\subsection{Latent Action Modeling for VLA Pretraining}

Generalist vision-language-action (VLA) models, such as OpenVLA~\cite{Kim2024OpenVLAAO}, OpenVLA-OFT~\cite{kim2025fine}, $\pi_0$~\cite{Black20240AV}, $\pi_{0.5}$~\cite{intelligence2025pi05}, and DexVLA~\cite{Wen2025DexVLAVM}, have demonstrated strong multi-task and multi-robot manipulation capabilities through large-scale robot co-training. 

Latent action modeling provides an alternative interface for learning action-relevant representations from videos without explicit control labels. Genie~\cite{bruce2024genie} demonstrates that discrete latent actions can capture controllable dynamics in videos, while LAPA~\cite{Ye2024LatentAP}, UniVLA~\cite{Bu2025UniVLALT}, Villa-X~\cite{chen2026villa}, HARP-VLA~\cite{zhu2026harp} extend this paradigm to robot learning and VLA pretraining by using latent motion representations to bridge action-free videos and downstream continuous control. 
Despite these advances, most LAMs remain driven by RGB transitions and provide limited explicit supervision for fine-grained end-effector articulation. Recent motion-aware LAMs~\cite{chen2026villa, Dai2026ConLACL, zhu2026harp} improve physical grounding, but often focus on predefined motion representations rather than shared fine-grained geometry across embodiments.


\subsection{Cross-Embodiment Imitation and Human-Robot Transfer}

Prior work addresses embodiment gaps in appearance, morphology, and action spaces through transferable representations and cross-embodiment imitation.
XSkill~\cite{Xu2023XSkillCE}, UniSkill~\cite{Kim2025UniSkillIH}, and EgoMimic~\cite{Kareer2024EgoMimicSI} learn transferable representations from human and robot demonstrations, while Human2Robot~\cite{Xie2025Human2RobotLR} and Zhu et al.~\cite{zhu2025learning} explore transferring manipulation knowledge from human videos to robot policies.
Recent approaches exploit geometric abstractions to bridge embodiment differences. LAD~\cite{Bauer2025LatentAD} learns cross-embodiment latents from retargeted paired poses with contrastive alignment; OPFA~\cite{Mu2026OnePolicyFitsAllGA} derives point-cloud representations from robot joint states and forward kinematics; XL-VLA~\cite{Jiang2026CrossHandLR} aligns dexterous hands through forward-kinematics-constrained shared latents; and METIS~\cite{Fu2025METISME} uses unified wrist or fingertip trajectories and 3D hand-motion tokens for VLA supervision. 
In contrast, GALA learns transition-level latent actions from end-effector geometry without cross-embodiment point, joint, or semantic-keypoint correspondence, while preserving native action spaces.

\subsection{3D Hand and Point-Cloud Representations}

MANO~\cite{Romero2017EmbodiedH} provides a standard parametric model of the human hand, while monocular reconstruction methods such as HaMeR~\cite{Pavlakos2023ReconstructingHI} and WiLoR~\cite{Potamias2024WiLoRE3} enable 3D hand geometry to be recovered from RGB videos.
Meanwhile, point-based architectures, including Point-BERT~\cite{Yu2021PointBERTP3}, Point-MAE~\cite{Pang2022MaskedAF}, and Point Transformer V3~\cite{Wu2023PointTV}, have demonstrated strong capability in learning transferable 3D geometric features.

Unlike joint vectors, fixed-topology meshes, or semantic keypoints, unordered point clouds do not require different end effectors to share the same kinematic structure or topology.
They therefore provide a common geometric representation for human hands, dexterous robot hands, simulated hands, and parallel-jaw grippers.
However, a common representation is not necessarily embodiment-invariant, as point-cloud geometry can still encode embodiment-specific differences in shape and scale.

\section{Method}
GALA consists of two stages: geometry-aware latent-action learning and geometry-aware VLA co-training. 
In the first stage, human and robot end effectors are represented as wrist- or root-centered point clouds under a unified format. Given a start--goal interval, the model jointly learns visual and geometric latent actions from RGB and end-effector geometry transitions. Our Unified End-effector Motion Representation (UEMR) shares the geometric encoder, codebook, and decoder across embodiments, while conditioning goal reconstruction on the initial hand geometry, encouraging the latent codes to capture transferable motion semantics without requiring joint correspondence or a unified action space. 
In the second stage, the visual and geometric latent codes supervise VLA co-training through dedicated bridge tokens, while a shared flow-matching action expert predicts continuous actions with embodiment-specific action heads.

\subsection{Geometry-Aware Latent Action Model}

Our goal is to capture fine-grained end-effector motions across heterogeneous embodiments without relying on a unified action space.
Since image-based latent actions may overlook subtle finger articulation and contact changes, we introduce a discrete \emph{geometric latent action} alongside the visual latent action, using 3D end-effector geometry transitions as structured motion supervision.

\begin{figure*}[h]
    \centering
    \includegraphics[width=0.98\linewidth]{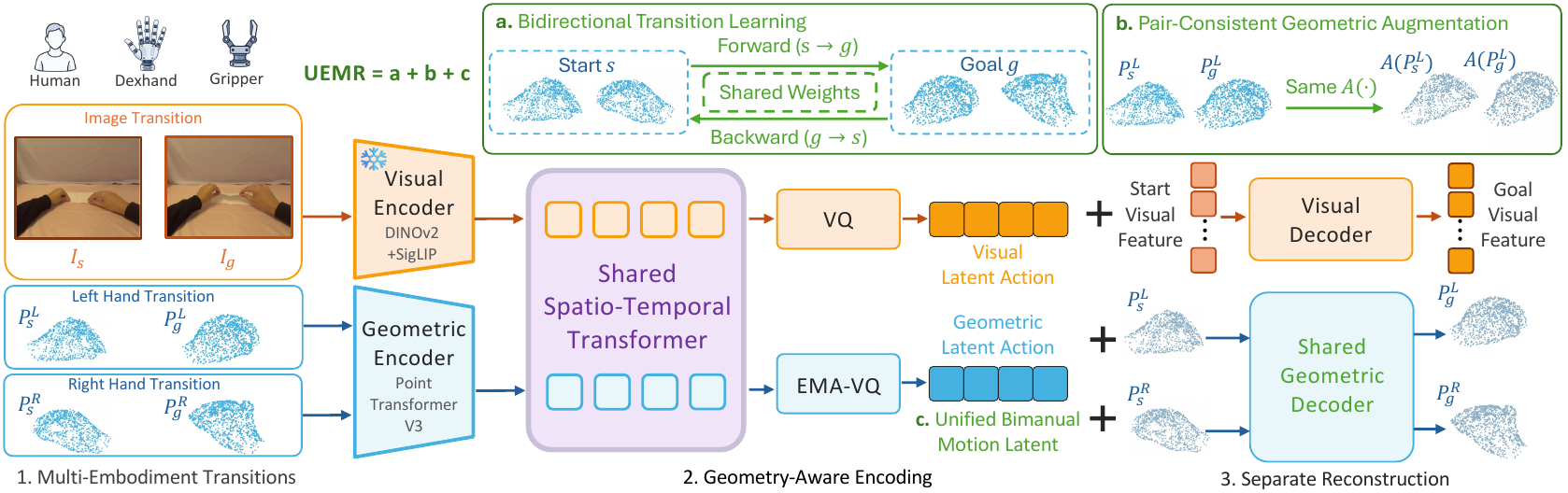}
    \vspace*{-3mm}
    \captionof{figure}{\textbf{Geometry-aware latent-action learning}. GALA jointly learns visual and geometric latent actions, with UEMR encouraging geometric latents to represent transferable end-effector transitions across embodiments.}
    \label{fig:stage1}
    \vspace*{-6mm}
\end{figure*}

\subsubsection{Point Cloud Generation}

For embodiment $e$, hand side $h\in{L,R}$, and time $t$, we represent the end-effector geometry as an unordered point cloud
\begin{equation}
P_t^{e,h}
=
{p_{t,i}^{e,h}}_{i=1}^{N}
\in
\mathbb{R}^{N\times 3},
\qquad N=1024.
\label{eq:point_cloud_definition}
\end{equation}
All point clouds are expressed in a wrist- or root-centered local frame under a common axis convention and are sampled from the corresponding surface geometry. This representation requires no point, joint, or mesh-topology correspondence across embodiments.

\noindent\textbf{Human Hands.}
For each RGB frame, WiLoR~\cite{Potamias2024WiLoRE3} estimates the left and right MANO~\cite{Romero2017EmbodiedH} meshes and hand keypoints. We transform each detected hand into a palm-aligned local frame centered at the wrist and sample its MANO surface to obtain the corresponding point cloud.

\noindent\textbf{Robot Embodiments.}
For robot embodiments, we obtain the end-effector geometry from the corresponding URDF or MJCF model. Given the recorded joint configuration, forward kinematics transforms surface samples from the end-effector links into their current poses, which are then aggregated and resampled to $N$ points. The same procedure applies to both dexterous hands and parallel-jaw grippers.
Further details of coordinate-frame construction, validity handling, and embodiment-specific state conversion are provided in the Appendix.

\subsubsection{Visual and Geometric Action Encoding}

As shown in Fig.~\ref{fig:stage1}, given a start--goal interval $(s,g)$, the model takes RGB observations $(I_s,I_g)$, a language instruction $\ell$, and the corresponding left- and right-hand point-cloud pairs $(P_s^L,P_g^L)$ and $(P_s^R,P_g^R)$.
The start--goal point clouds are normalized using a shared center and scale to reduce cross-dataset scale variation while preserving their relative geometric motion.
Missing hands are excluded using validity masks.

Frozen DINOv2~\cite{oquab2023dinov2} and SigLIP~\cite{zhai2023sigmoid}
encode the RGB observations, while a frozen Point Transformer V3 (PTv3)~\cite{Wu2023PointTV} encodes the hand point clouds.
The resulting visual and geometric features are fused with the language condition through a shared spatiotemporal Transformer.
The model produces two complementary latent-action representations: visual tokens that capture scene-level dynamics and shared geometric latent tokens that capture the bimanual 3D geometric transition.

Let
\[
Z^G=\{z_j^G\}_{j=1}^{N_p}
\]
denote the geometric latent representation, where $N_p$ denotes the number of point-cloud codes.
Each geometric token is quantized using a shared EMA-VQ codebook $\{e_k\}_{k=1}^{K_G}$, where $K_G$ denotes codebook size:
\begin{equation}
k_j
=
\arg\min_k
\left\|z_j^G-e_k\right\|_2^2,
\qquad
q_j^G=e_{k_j}.
\label{eq:hand_vq}
\end{equation}

The geometric encoder and codebook are shared across human hands, dexterous robot hands, and parallel-jaw grippers, providing a common discrete latent space for heterogeneous end-effector transitions.

\subsubsection{Unified End-effector Motion Representation}

To model end-effector motion across heterogeneous morphologies, while improving cross-embodiment generalizability, we introduce the \textbf{U}nified \textbf{E}nd-effector \textbf{M}otion \textbf{R}epresentation (UEMR) tailored to GALA, which includes three core designs: unified bimanual motion latent, pair-consistent geometric augmentation, and bidirectional transition learning. 

\noindent\textbf{Unified Bimanual Motion Latent.}
The geometric latent tokens jointly represent the bimanual geometric transition over the sampled interval rather than being assigned separately to the left and right hands. 
As for single-arm gripper datasets, we use the same point cloud as input to both branches.
Empirically, this design facilitates the joint modeling of point-cloud latent actions across different numbers of arms.
The same quantized geometric latent representation is combined with each hand's initial geometry, and a shared geometric decoder reconstructs the corresponding goal geometry:
\begin{equation}
\hat{P}_g^R
=
D_G
\left(
P_s^R,
Q(Z^G)
\right),
\hat{P}_g^L
=
D_G
\left(
P_s^L,
Q(Z^G)
\right).
\label{eq:uemr_reconstruction}
\end{equation}

The decoder predicts the normalized goal point clouds and is supervised with the Chamfer Distance (CD):
\begin{equation}
\mathcal{L}_{G\text{-rec}}
=
m^R
\mathrm{CD}
\left(
\hat{P}_g^R,
P_g^R
\right)
+
m^L
\mathrm{CD}
\left(
\hat{P}_g^L,
P_g^L
\right).
\label{eq:hand_reconstruction_loss}
\end{equation}
where $m^R$ and $m^L$ denote the validity masks for the right and left hands, respectively.


Conditioning the decoder on the initial geometry provides embodiment-specific geometric context, encouraging the shared latent to encode the start--goal geometric transition rather than static morphology.

\noindent\textbf{Pair-Consistent Geometric Augmentation.}
To reduce sensitivity to dataset-specific coordinates while preserving the relative geometric transition, we apply a pair-consistent random 3D transformation to each start--goal point-cloud pair:
\begin{equation}
\mathcal{A}(P)
=
aRP+t,
(P_s^h,P_g^h)
\mapsto
\left(
\mathcal{A}(P_s^h),
\mathcal{A}(P_g^h)
\right).
\label{eq:point_cloud_augmentation}
\end{equation}
where $a$, $R$, and $t$ denote the sampled scale, rotation, and translation, respectively. Applying the same transformation to both endpoints preserves their relative geometric motion while perturbing the absolute coordinate frame.

\noindent\textbf{Bidirectional Transition Learning.}
To further encourage the latent representation to capture relative temporal dynamics, we train the latent-action model in both temporal directions. 
Let $X_t=(I_t,P_t^L,P_t^R)$ denote the multimodal observation at time $t$. 
For each hand $h\in\{L,R\}$, the forward and backward transitions are modeled as
\begin{align} Z_f^G &= E_G(X_s,X_g,\ell), \hat{P}_g^h = D_G\!\left(P_s^h,Q(Z_f^G)\right), \nonumber\\ Z_b^G &= E_G(X_g,X_s,\ell), \hat{P}_s^h = D_G\!\left(P_g^h,Q(Z_b^G)\right). \label{eq:bidirectional_hand_reconstruction} \end{align}
Both directions share the same encoder, codebook, and decoder. The backward objective provides additional transition supervision without imposing an explicit relationship between $Z_f^G$ and $Z_b^G$.

For each temporal direction, we jointly optimize the visual and geometry-aware latent-action objectives:
\begin{equation}
\begin{aligned}
\mathcal{L}_{forward}
&=
\mathcal{L}_{I\text{-rec}}
+\mathcal{L}_{I\text{-codebook}}
+\beta\mathcal{L}_{I\text{-commit}}
\\
&\quad
+\lambda_G\mathcal{L}_{G\text{-rec}}
+\beta\lambda_G\mathcal{L}_{G\text{-commit}}.
\end{aligned}
\label{eq:forward_lam_loss}
\end{equation}
$\mathcal{L}_{I\text{-rec}}$, $\mathcal{L}_{I\text{-codebook}}$ and $\mathcal{L}_{I\text{-commit}}$ supervise the visual latent-action branch in the conventional VQ manner. 
$\mathcal{L}_{G\text{-rec}}$ supervises the UEMR geometric reconstruction, while $\mathcal{L}_{G\text{-commit}}$ regularizes the discrete geometric latent in EMA-VQ manner.

The full latent-action objective combines the forward and backward directions:
\begin{equation}
\mathcal{L}_{\mathrm{LAM}}
=
\mathcal{L}_{forward}
+
\mathcal{L}_{backward}.
\label{eq:lam_loss}
\end{equation}

Together, unified bimanual motion latent, pair-consistent augmentation, and bidirectional training encourage UEMR to capture more generalizable end-effector motion rather than static morphology, absolute pose, or dataset-specific coordinate shortcuts. The resulting visual and geometric latent-action codes are used as supervision for downstream VLA co-training.





\begin{figure*}[h]
    \centering
    \includegraphics[width=0.7\linewidth]{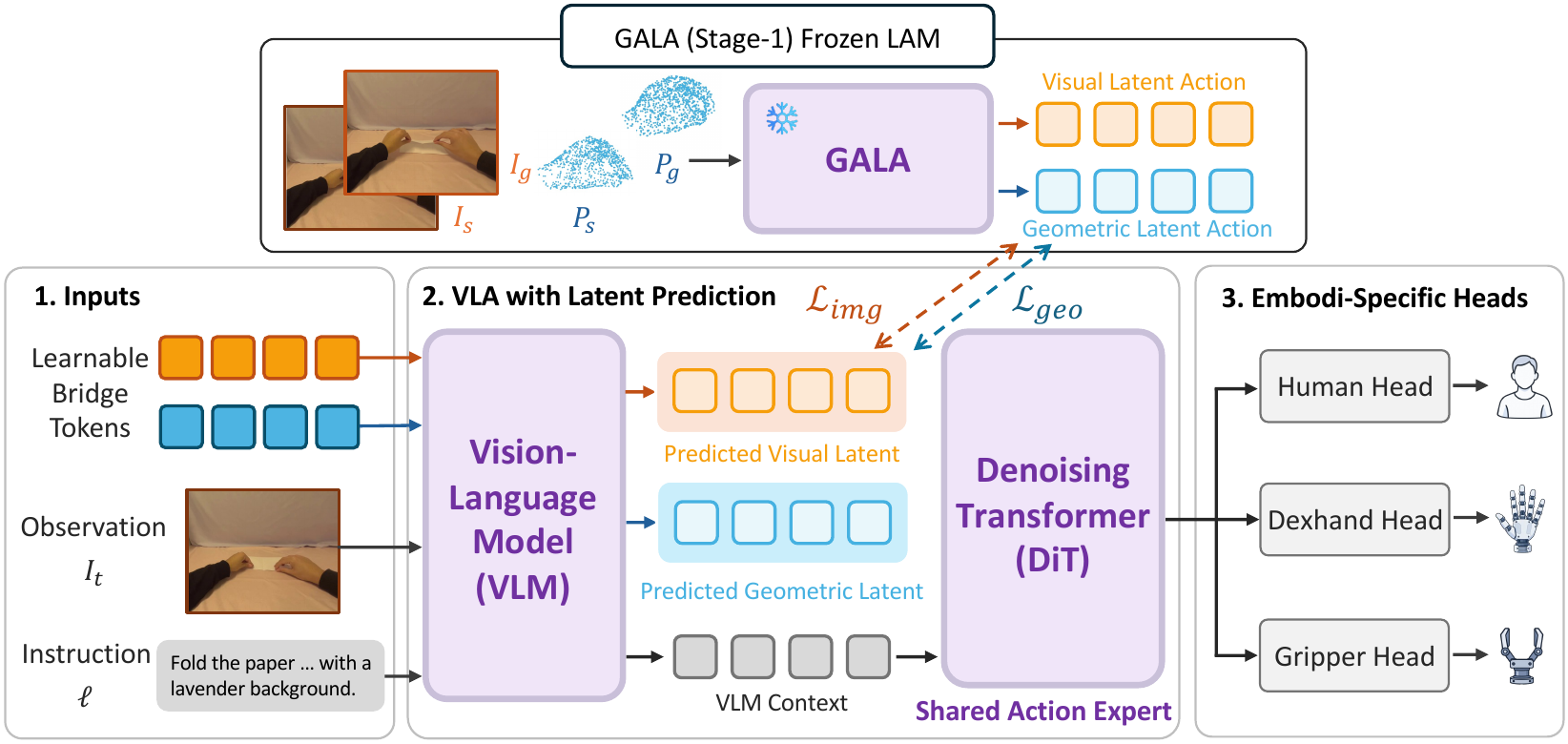}
    \vspace*{-3mm}
    \captionof{figure}{\textbf{Geometry-aware VLA co-training.} Frozen GALA latents supervise dedicated VLM bridge tokens, while a shared action expert is grounded into embodiment-specific native action spaces.}
    \label{fig:stage2}
    \vspace*{-6mm}
\end{figure*}

\subsection{Geometry-Aware VLA Co-Training}

We build our policy upon the GR00T architecture~\cite{bjorck2025gr00t}, which consists of a vision--language backbone for multimodal reasoning and a flow-matching action expert for continuous action generation. 
As shown in Fig.~\ref{fig:stage2}, given the current observation $o_t$, embodiment state $s_t$, and language instruction $\ell$, the policy predicts an action chunk $A_t=[a_t,\ldots,a_{t+H-1}]$. To inject structured motion and geometric supervision into the vision--language representation, we incorporate the Geometry-Aware Latent Action Model introduced above and jointly optimize latent-action prediction and continuous action generation.

\subsubsection{Geometry-Aware Latent Action Prediction}
The Geometry-Aware Latent Action Model produces two complementary discrete latent-action sequences: a visual latent-action sequence
$k_t^{\mathrm{img}}=[k_{t,1}^{\mathrm{img}},\ldots,k_{t,N_v}^{\mathrm{img}}]$
and a geometric latent action sequence
$k_t^{\mathrm{geo}}=[k_{t,1}^{\mathrm{geo}},\ldots,k_{t,N_p}^{\mathrm{geo}}]$.
We append two groups of learnable bridge tokens,
$q^{\mathrm{img}}$ and $q^{\mathrm{geo}}$, to the VLM input sequence. Conditioned on the current observation and language instruction, the hidden states associated with these bridge tokens are projected onto their respective latent-action vocabularies to autoregressively predict the visual and geometric latent-action tokens:
\begin{equation}
\begin{gathered}
    \hat{k}^{\mathrm{img}}_t,\hat{k}^{\mathrm{geo}}_t
    =
    f_{\mathrm{VLM}}
    \left(
        I_t,\ell,
        q^{\mathrm{img}},q^{\mathrm{geo}}
    \right),
\end{gathered}
\end{equation}
\begin{equation}
\begin{gathered}
    \begin{aligned}
        \mathcal{L}_{\mathrm{latent}}
        &=
        \mathcal{L}_{\mathrm{img}}
        + \lambda_{\mathrm{geo}}\mathcal{L}_{\mathrm{geo}} \\
        &=
        \mathrm{CE}\!\left(
            k^{\mathrm{img}}_t,\hat{k}^{\mathrm{img}}_t
        \right)
        +
        \lambda_{\mathrm{ge o}}\,
        \mathrm{CE}\!\left(
            k^{\mathrm{geo}}_t,\hat{k}^{\mathrm{geo}}_t
        \right).
    \end{aligned}
\end{gathered}
\end{equation}
where $\mathcal{L}_{\mathrm{img}}$ and $\mathcal{L}_{\mathrm{geo}}$ denote the cross-entropy losses against the visual and geometric latent-action codes, respectively. 
The two prediction targets provide complementary supervision: visual latent actions capture appearance-level interaction dynamics, whereas geometric latent actions explicitly encode the underlying 3D geometric transitions. Their joint prediction encourages the VLM to learn representations that capture both semantic task intent and geometry-aware physical dynamics.

\subsubsection{Real Action Representation}
Rather than manually projecting heterogeneous embodiments into a unified action space, we retain the native action representation of each embodiment. For human-hand demonstrations, we define the human action as the 3D positions and rotations of both wrists expressed in the camera coordinate frame, together with the positions of the finger keypoints expressed in their corresponding wrist coordinate frames. This representation decomposes human-hand motion into the global movement of each hand and its local finger articulation. For robotic embodiments, including parallel-jaw grippers and dexterous hands, we directly adopt the action space originally defined by each dataset, thereby preserving its native control semantics and parameterization.

\subsubsection{Shared DiT with Embodiment-Specific Action Heads}
To accommodate the resulting differences in action dimensionality and semantics, the action expert consists of a shared diffusion transformer (DiT)~\cite{peebles2023scalable} and a lightweight embodiment-specific action head $h_e$ for each embodiment $e$. The shared DiT models transferable visuomotor dynamics across embodiments, while each action head maps the shared representation to the corresponding native action space $\mathcal{A}_e$. Specifically, the shared DiT processes the VLM context $x_t$, current embodiment state $s_t$, noisy action chunk $A_t^\tau$, and flow timestep $\tau$:
\begin{equation}
    V_{\theta,e}
    \left(
        A_t^\tau
        \mid x_t,s_t,\tau
    \right)
    =
    h_e\left(
        f_{\mathrm{DiT}}
        \left(
            A_t^\tau,x_t,s_t,\tau
        \right)
    \right),
\end{equation}
where $A_t^\tau=\tau A_t+(1-\tau)\epsilon, \tau\sim\mathcal{U}[0,1], \epsilon\sim\mathcal{N}(\mathbf{0},\mathbf{I})$.
For each training sample from embodiment $e$, only its corresponding action head is activated and optimized. The embodiment-specific flow-matching objective is defined as
\begin{equation}
    \mathcal{L}_{\mathrm{action}}
    = 
    \mathbb{E}_{\tau,\epsilon}
    \left[
        \left\|
        V_{\theta,e}
        \left(
            A_t^\tau\mid x_t,s_t,\tau
        \right)
        -
        \left(A_t-\epsilon\right)
        \right\|_2^2
    \right].
\end{equation}
This design shares high-level action reasoning and temporal dynamics across embodiments through the common DiT, while preserving the dimensionality and control semantics of each action space through separate output heads.

The overall VLA co-training objective combines visual latent-action prediction, geometric latent-action prediction, and embodiment-specific continuous action generation:
\begin{equation}
    \mathcal{L}_{\mathrm{VLA}}
    =
    \mathcal{L}_{\mathrm{action}} + 
    \lambda_{\mathrm{latent}}\mathcal{L}_{\mathrm{latent}}.
\end{equation}
Through this joint objective, both human and robot demonstrations contribute to learning geometry-aware physical representations, while the shared DiT extracts transferable visuomotor dynamics from heterogeneous embodiments. Meanwhile, the embodiment-specific action heads ground the shared representation into executable controls without requiring an artificially unified action space.

\section{Experiment}

We evaluate GALA from three perspectives. 

\textbf{Q1: Can GALA model fine-grained end-effector motions?}
We perform probing evaluation to measure whether the frozen GALA latent action captures fine-grained motion information required for dexterous actions (\ref{sec:fine_grain}).

\textbf{Q2: Does GALA learn latent actions with shared semantics across embodiments?}
We analyze the learned latent action codebook and evaluate the cross-embodiment semantic alignment through a retrieval task.
 (\ref{sec:retrieval}).

\textbf{Q3: Can GALA benefit downstream VLA learning and embodiment scaling?}
We integrate GALA into downstream VLA training and evaluate whether its representations improve policy performance (\ref{sec:robocasa},~\ref{sec:real_world}).

\noindent\textbf{Dataset Components.}
In both Stage 1 and Stage 2, we train GALA and the subsequent VLA using data from four embodiments: human hand, Fourier hand, ROBOTERA XHand, and the Robotiq gripper. Our training data comprise EgoDex~\cite{hoque2026egodex}, HOI4D~\cite{liu2022hoi4d}, DROID~\cite{khazatsky2024droid}, RoboCasa-GR1~\cite{bjorck2025gr00t,Nasiriany2024RoboCasaLS}, as well as self-collected XHand and human-hand manipulation datasets.

\noindent\textbf{Baselines and Protocol.}
We compare several LAM variants and baselines against GALA. Different experiments use different subsets of these baselines according to their applicable data modalities and evaluation settings.
\begin{itemize}
\item \textbf{GALA w/o PC} removes the hand point-cloud branch and learns latent actions solely from image pairs.
\item \textbf{GALA w/o UEMR} removes three UEMR designs, using seperate latent codes for left and right hands.
\item \textbf{METIS}~\cite{Fu2025METISME} uses sparse hand representation based on 6D wrist poses and 6D fingertip features.
\item \textbf{Native Kinematics} encodes embodiment-specific joint states or end-effector kinematics into latent actions using the corresponding definitions or  annotations.
\item \textbf{OPFA}~\cite{Mu2026OnePolicyFitsAllGA} applies the OPFA geometry encoder to the same point-cloud inputs used by GALA and concatenates the start and goal hand-state latents to form a transition representation.
\item \textbf{UniVLA}~\cite{Bu2025UniVLALT} follows the original LAM framework and training of UniVLA, while using the same downstream VLA architecture and supervision as GALA.
\end{itemize}
GALA, GALA w/o PC, GALA w/o UEMR, METIS and UniVLA are trained on the same human-and-robot video dataset, whereas Native Kinematics and OPFA use only robot data.
Across methods, we keep the episode, start--goal frame pair, prediction target, and probe train/validation/test splits identical.


\subsection{Fine-Grained Motion Probing Evaluation}
\label{sec:fine_grain}

To evaluate the fine-grained motion information preserved in different latent-action representations, we freeze the latent-action encoders, VQ codebooks, and feature backbones, and train a lightweight MLP to predict robot motion. For each start--goal pair, the probe predicts
\begin{equation}
    y =
    \left[
    \Delta p,\,
    \Delta R,\,
    \Delta q_{\mathrm{finger}}
    \right],
    \label{eq:motion_probe_target}
\end{equation}
where $\Delta p$ denotes the end-effector translation, $\Delta R$ denotes the wrist rotation, and $\Delta q_{\mathrm{finger}}$ denotes the finger or gripper articulation.

We adopt a unified \textbf{State+Image+Hand} probe setting, where the MLP takes the current robot state, the frozen visual latent, and the geometric latent representation of the corresponding method as input. For \textbf{GALA w/o PC}, which does not contain a geometric latent branch, the probe uses only State+Image as input. All methods use the same MLP architecture and training settings; detailed network configurations and hyperparameters are provided in the Appendix. All quantitative probes are evaluated on the same held-out test set split from the XHand dataset, and the reported results are averaged over three fixed random seeds.

We report \textbf{Position Error}, \textbf{Rotation Error}, and \textbf{Finger Error}. Position Error is defined as the Euclidean error between the predicted and target end-effector displacements, measured in centimeters. Rotation Error is defined as the geodesic angular error between the predicted and target wrist rotations, measured in degrees. Finger Error is defined as the mean angular error over all valid finger or gripper joints, also measured in degrees.

As shown in Table~\ref{tab:representation_probe}, GALA achieves the best performance across all fine-grained motion prediction metrics, indicating that its latent representation preserves more fine-grained end-effector motion information.
\begin{table}[t]
    \centering
    \caption{Fine-grained motion probe results. }
    \label{tab:representation_probe}
    \small
    \setlength{\tabcolsep}{5pt}
    \renewcommand{\arraystretch}{1.15}
    \begin{tabular}{lccc}
        \hline
        Method
        & Pos. $\downarrow$
        & Rot. $\downarrow$
        & Finger $\downarrow$ \\
        \hline
        METIS
        & 6.162 & 8.520 & 14.328 \\
        Native Kinematics
        & 5.879 & 8.098 & 7.768 \\
        OPFA
        & 6.102 & 8.260 & 7.707 \\
        GALA w/o PC
        & 6.790 & 8.567 & 15.113 \\
        \textbf{GALA (Ours)}
        & \textbf{5.359}
        & \textbf{7.898}
        & \textbf{7.585} \\
        \hline
    \end{tabular}
\end{table}



\subsection{Cross-Embodiment Latent Action Retrieval}
\label{sec:retrieval}
We evaluate whether the learned latent actions capture motion semantics shared across embodiments through a cross-embodiment retrieval benchmark. We annotate local \emph{grasp}, \emph{hold}, and \emph{release} transitions from four embodiments: a human hand, a parallel-jaw gripper, and two dexterous robot hands. 
For each transition, we extract the frozen post-quantization representation from the annotated start–goal pair. 
The resulting representations are flattened, $\ell_2$-normalized, and compared using cosine similarity.
Retrieval is performed strictly across embodiments. Each query from embodiment $e_q$ is independently retrieved against every other embodiment $e_g$, where $e_q \neq e_g$, and gallery samples sharing the query motion label are treated as positives. 
We evaluate two tracks: a full human--robot track comparing GALA, GALA w/o PC, GALA w/o UEMR, and METIS, and a robot-only track that additionally includes Native Kinematics and OPFA.
We report R@1, indicating whether the nearest retrieved sample matches the query motion; P@5, measuring the proportion of motion-matched samples among the top five results; and mAP, evaluating the ranking quality of all positive samples over the full gallery.Metrics are macro-averaged first across the three motion classes and then across all directed embodiment pairs.
As shown in Table II, GALA consistently achieves the best cross-embodiment retrieval performance in both human–robot and robot-only settings, demonstrating stronger preservation of shared motion semantics across heterogeneous embodiments.

\begin{table}[t]
    \centering
    \small
    \setlength{\tabcolsep}{4.5pt}
    \caption{
        Cross-embodiment motion retrieval results (\%).
    }
    \label{tab:cross_emb_retrieval}
    \begin{tabular}{lccc}
        \toprule
        Method
        & R@1 $\uparrow$
        & P@5 $\uparrow$
        & mAP $\uparrow$ \\
        \midrule

        \multicolumn{4}{l}{\textit{Full human--robot track}} \\
        METIS
        & 33.26
        & 33.29
        & 37.21 \\
        GALA w/o PC
        & {33.37}
        & {33.37}
        & 36.38 \\
        GALA w/o UEMR
        & 38.69
        & 40.31
        & 48.15 \\
        \textbf{GALA (Ours)}
        & \textbf{45.67}
        & \textbf{44.57}
        & \textbf{49.49} \\
        \midrule

        \multicolumn{4}{l}{\textit{Robot-only track}} \\
        METIS
        & 33.30
        & 33.30
        & 36.10 \\
        Native Kinematics
        & 39.31
        & 35.46
        & 41.94 \\
        OPFA
        & {41.18}
        & {36.83}
        & {43.03} \\
        GALA w/o PC
        & 33.38
        & 33.38
        & 36.41 \\
        GALA w/o UEMR
        & 33.83
        & 38.54
        & 43.11 \\
        \textbf{GALA (Ours)}
        & \textbf{46.91}
        & \textbf{41.74}
        & \textbf{46.95} \\
        \bottomrule
    \end{tabular}
\end{table}

\subsection{RoboCasa Simulation Experiments}
\label{sec:robocasa}
We evaluate GALA on the RoboCasa GR-1 Tabletop benchmark, a dexterous manipulation environment for the GR-1 humanoid robot equipped with dual multi-finger hands.
The benchmark contains 24 tasks, including placing objects into drawers, microwaves, and cabinets followed by closing them, as well as transferring novel objects from cutting boards, placemats, plates, and trays to diverse receptacles.
We report the average success rate across all tasks.

We consider two experimental settings.
First, we conduct a controlled experiment using only GR-1 data to isolate the effect of latent action modeling.
All methods use the same VLA architecture, training data, batch size, and optimization steps, differing only in their latent action models.
We compare GALA with UniVLA, METIS, OPFA, and native-kinematics supervision, together with GALA variants that remove point-cloud modeling or UEMR.

\begin{table}[t]
\centering
\caption{
Controlled comparison on RoboCasa GR-1 using only GR-1 data.
All methods differ only in their latent action models.
}
\label{tab:robocasa_gr1_ablation}
\resizebox{0.65\linewidth}{!}{
\begin{tabular}{lc}
\toprule
LAM Method & Success Rate (\%) $\uparrow$ \\
\midrule
UniVLA              & 48.0 \\
METIS               & 43.8 \\
Native Kinematics   & 51.8 \\
OPFA                & 53.5 \\
\midrule
GALA w/o PC         & 50.9 \\
GALA w/o UEMR       & 54.4 \\
\textbf{GALA (Ours)}       & \textbf{55.7} \\
\bottomrule
\end{tabular}
}
\end{table}

As shown in Table~\ref{tab:robocasa_gr1_ablation}, GALA achieves the highest success rate of 55.7\%.
Removing point-cloud modeling reduces performance to 50.9\%, and removing UEMR reduces performance to 54.4\%, demonstrating the importance of modeling semantically shared end-effector geometry motion beyond image-level displacement.
GALA also outperforms native-kinematics supervision, suggesting that its learned geometric representation provides more effective supervision for downstream VLA training.

Second, we jointly pretrain VLA model on Fourier-hand, XHand, human-hand, and Robotiq gripper data to evaluate the scalability of GALA across heterogeneous embodiments. UniVLA, GALA w/o UEMR, and GALA use the same multi-embodiment training settings.
The resulting model is evaluated on the same 24 GR-1 tasks and compared with state-of-the-art methods.

\begin{table}[t]
\centering
\caption{
Comparison on RoboCasa GR-1 under multi-embodiment co-training.
GR-1-only denotes GR-1 data only training.
}
\label{tab:robocasa_gr1_cotrain}
\resizebox{1.0\linewidth}{!}{
\begin{tabular}{lcc}
\toprule
Method & Success Rate (\%) $\uparrow$ & Comp. w/ GR-1-only $\uparrow$ \\
\midrule
FLARE~\cite{zheng2025flare} & 55.0 & - \\
DiT4DiT~\cite{ma2026dit4dit} & 56.7 & - \\
JoyAI-RA~\cite{zhang2026joyai} & 63.2 & - \\
UniT~\cite{chen2026unit} & 66.8 & - \\
UniVLA & 53.6 & +5.6 \\
\midrule
GALA w/o UEMR   & 58.6  & +4.2 \\ 
\textbf{GALA (Ours)}   & \textbf{68.3} & \textbf{+12.6} \\
\bottomrule
\end{tabular}
}
\end{table}

As shown in Table~\ref{tab:robocasa_gr1_cotrain}, GALA achieves a success rate of 68.3\%, outperforming JoyAI-RA, UniT, UniVLA and other publicly reported baselines.
The highest success rate improvement over the GR-1-only result $+12.6\%$ further indicates that UEMR enables GALA to transfer complementary manipulation knowledge across heterogeneous embodiments, offering a promising foundation for scaling generalist VLA towards increasingly diverse embodiments.

\subsection{Real-World Robot Experiments}
\label{sec:real_world}
We further evaluate GALA on four real-world manipulation tasks with different motion and precision requirements using a 12-DoF XHand robotic hand: \emph{Pick and Place}, where the robot grasps an object and places it at a target location; \emph{Push Box}, where it pushes a box toward a designated region; \emph{Press Button}, which requires precise finger positioning and contact; and \emph{Flip Cup}, which involves dexterous contact and orientation changes.
For a fair comparison, all methods use the same amount of real-world training data and differ in backbone architecture, evaluated over 50 trials per task, with further details in the Appendix.

\begin{table}[t]
\centering
\caption{
Success rates (\%) on four real-world manipulation tasks.
}
\label{tab:realworld_results}
\resizebox{\linewidth}{!}{
\begin{tabular}{lccccc}
\toprule
Method & Pick & Push & Press & Flip & Average \\
\midrule
OpenVLA          & 0  & 24 & 18 & 10 & 13.0 \\
UniVLA           & 38 & 62 & 32 & 20 & 38.0 \\
OpenVLA-OFT      & 52 & 72 & 54 & 42 & 55.0 \\
$\pi_0$          & 56 & 72 & 56 & 34 & 54.5 \\
$\pi_{0.5}$      & 72 & 80 & 68 & 52 & 68.0 \\
HARP-VLA         & 72 & \textbf{82} & 74 & 58 & 71.5 \\
\midrule
GALA w/o UEMR    & 74 & 78 & 72 & 50 & 68.5 \\
\textbf{GALA (Ours)}    & \textbf{80} & \textbf{82} & \textbf{78} & \textbf{62} & \textbf{75.5} \\
\bottomrule
\end{tabular}
}
\end{table}

As shown in Table~\ref{tab:realworld_results}, GALA achieves the best average success rate of 75.5\%, outperforming HARP-VLA by 4.0 points and $\pi_{0.5}$ by 7.5 points, demonstrating the effectiveness of our method for real-world dexterous manipulation. 
Removing UEMR reduces the average success rate by 7.0 points, demonstrating that UEMR improves the cross-embodiment generalizability of latent actions, thereby facilitating downstream VLA training.

\section{Conclusions}

In this work, we presented GALA, a geometry-aware latent action modeling framework for learning fine-grained manipulation representations from multi-embodiment data. 
To address the heterogeneity of end-effector morphologies, we introduced UEMR tailored to GALA, which preserves fine-grained motion information while improving the cross-embodiment generalizability of latent actions.
By jointly modeling image-based motion features and end-effector point-cloud dynamics, GALA provides fine-grained supervision for transferring dexterous manipulation knowledge across embodiments, including unlabeled egocentric human videos. 
Experiments on fine-grained motion probing, cross-embodiment retrieval, and downstream VLA evaluation in RoboCasa and real-world settings demonstrate GALA’s effectiveness in modeling generalizable fine-grained actions across embodiments.
We believe that geometry-aware latent actions offer a promising foundation for scaling generalist VLAs toward increasingly diverse embodiments and dexterous manipulation tasks.








\small{
\bibliographystyle{IEEEtran}
\bibliography{references}
}

\end{document}